\documentclass[sigconf]{acmart}
\usepackage{cleveref}
\usepackage[gen]{eurosym}

\makeatletter
\newcommand\footnoteref[1]{\protected@xdef\@thefnmark{\ref{#1}}\@footnotemark}
\makeatother

\AtBeginDocument{%
  \providecommand\BibTeX{{%
    \normalfont B\kern-0.5em{\scshape i\kern-0.25em b}\kern-0.8em\TeX}}}

\copyrightyear{2020}
\acmYear{2020}
\setcopyright{rightsretained}
\acmConference[KDD '20]{Proceedings of the 26th ACM SIGKDD Conference on Knowledge Discovery and Data Mining}{August 23--27, 2020}{Virtual Event, CA, USA}
\acmBooktitle{Proceedings of the 26th ACM SIGKDD Conference on Knowledge Discovery and Data Mining (KDD '20), August 23--27, 2020, Virtual Event, CA, USA}
\acmDOI{10.1145/3394486.3403361}
\acmISBN{978-1-4503-7998-4/20/08}

\begin{document}
\fancyhead{}

\title{Interleaved Sequence RNNs for Fraud Detection}

\author{Bernardo Branco}
\authornote{The authors would like to thank Feedzai's Research Team, and especially Jacopo Bono, Hugo Ferreira, Jo\~ao Oliveirinha, and Marco Sampaio for valuable discussions, and help revising this document, and Miguel Almeida for work in a related prior project.}
\email{bernardo.branco@feedzai.com}
\affiliation{%
  \institution{Feedzai}
}

\author{Pedro Abreu}
\authornotemark[1]

\email{pedro.abreu@feedzai.com}
\affiliation{%
  \institution{Feedzai}
}

\author{Ana Sofia Gomes}
\authornotemark[1]
\email{sofia.gomes@feedzai.com}
\affiliation{%
 \institution{Feedzai}
}

\author{Mariana S. C. Almeida}
\authornotemark[0]
\authornotemark[1]
\authornote{Work developed while employed at Feedzai.}
\email{mariana@cleverly.ai}
\affiliation{\institution{Cleverly}}

\author{Jo\~ao Tiago Ascens\~ao}
\authornotemark[1]
\email{joao.ascensao@feedzai.com}
\affiliation{%
  \institution{Feedzai}
}

\author{Pedro Bizarro}
\authornotemark[1]
\email{pedro.bizarro@feedzai.com}
\affiliation{%
  \institution{Feedzai}
}

\begin{abstract}


Payment card fraud causes multibillion dollar losses for banks and merchants worldwide, often fueling complex criminal activities.
To address this, many real-time fraud detection systems use tree-based models, demanding complex feature engineering systems to efficiently enrich transactions with historical data while complying with millisecond-level latencies.
In this work, we do not require those expensive features by using recurrent neural networks and treating payments as an interleaved sequence, where the history of each card is an unbounded, irregular sub-sequence. We present a complete RNN framework to detect fraud in real-time, proposing an efficient ML pipeline from preprocessing to deployment.
We show that these feature-free, multi-sequence RNNs outperform state-of-the-art models 
saving millions of dollars in fraud detection and using fewer computational resources.

\end{abstract}



\settopmatter{printfolios=true}
\maketitle

\section{Introduction}
In payment card fraud, criminals execute physical or electronic payments without the cardholder's authorization. Banks and merchants are liable for accepting such transactions and have to reimburse cardholders. Credit card fraud alone was responsible for \$28B in losses in 2018 \cite{nilsonreport}.

Modern fraud detection systems combine rule-based systems with machine learning classification models to score transactions.
Frequently used models are tree-based learners, such as Random Forests~\cite{BHATTACHARYYA2011602}, XGBoost~\cite{Chen:2016:XST:2939672.2939785}, or LightGBM~\cite{NIPS2017_6907}. 
%
Tree-based learners rely heavily on data augmentation in the form of feature engineering to capture the usage patterns (or \emph{profiles}) of cardholders \cite{CORREABAHNSEN2016134, transactionaggregation}.
%
Profiles pose many challenges since they need to be refined during model development, and efficiently handled in production.

Deep Learning (DL) can produce state-of-the-art results in many areas without extensive feature engineering \cite{LeCun2015}. Recurrent Neural Networks (RNNs) such as Long Short-Term Memory (LSTM) \cite{doi:10.1162/neco.1997.9.8.1735} and Gated Recurrent Unit (GRU) \cite{ChoMGBSB14} are DL architectures addressing sequences and particularly suited for time-series data.
However, unlike in other domains, RNNs cannot be directly applied to fraud detection in production because the payment sequence is composed of multiple, interleaved, unbounded sequences, thereby requiring careful management of the internal RNN state.

In this paper, we use GRUs to build fraud detection models without intermediate feature engineering, suitable for mission-critical, streaming systems with millisecond latency requirements.

\begin{figure}[hb]
  \includegraphics[width=0.5\textwidth]{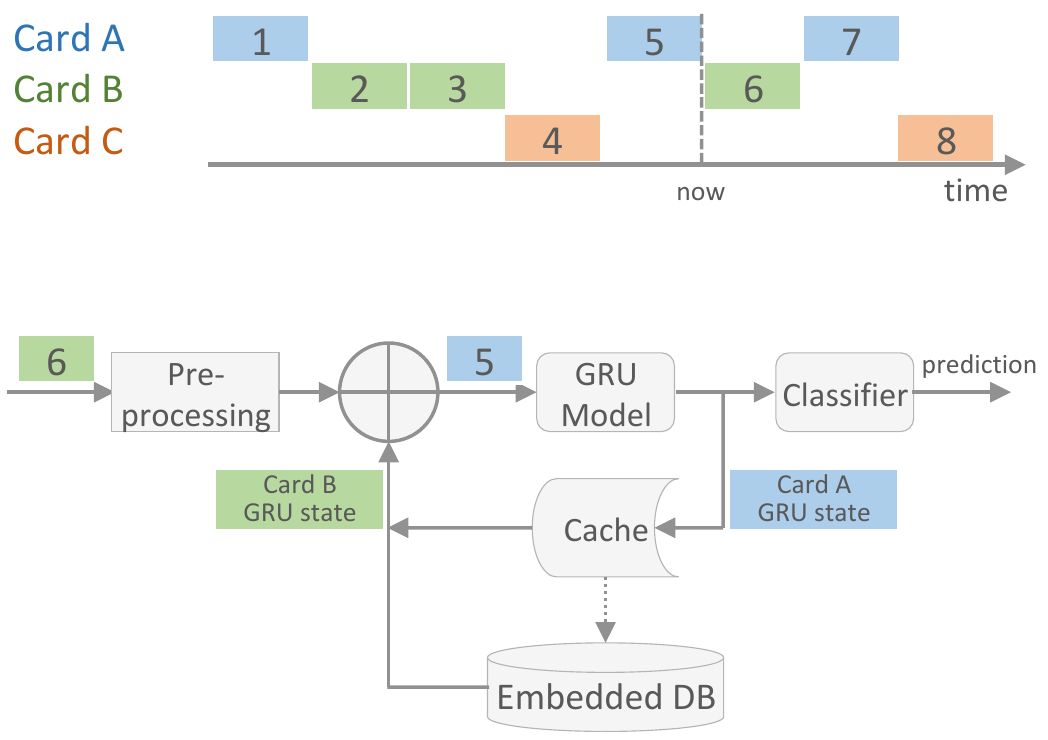}
  \caption{\emph{(top)} Interleaved transaction history of three cards. \emph{(bottom)} At run-time, the system joins transaction 6 with the previous Card B GRU state (last changed in transaction 3). Previously, it processed transaction 5 and stored the Card A GRU state in an embedded DB and memory cache.
}
  \label{fig:streaming-interleaved}
\end{figure}

We designed a system that is efficient during offline training and during online, streaming classification. Our main contributions are: 
\begin{enumerate}
    \item We identify a new type of problem: \emph{a sequence composed of many interleaved, unbounded sub-sequences} - i.e., the history of each card - with \emph{irregular time intervals between events} - i.e., transactions (section~\ref{subsec:modelarchitecture}).
    \item We propose an efficient batch training technique, sorting per sub-sequence and time, processing scorable and non-scorable events, limiting history, and chunking data to keep GPUs maximally occupied (sections~\ref{subsec:preprocessing}-\ref{subsec:batch_inference}).
    \item We introduce an efficient streaming inference technique, saving and restoring the GRU state, caching, and expiring events (section~\ref{subsec:streaming}).
    \item We evaluate the solution in two real-life use cases and show that these multi-sequence RNNs outperform state-of-the-art models in recall, but also latency, saving millions of dollars in fraud detection and resources (section~\ref{sec:validation}).
\end{enumerate}
Figure~\ref{fig:streaming-interleaved} depicts the workings and architecture of our system in a production scenario.

\section{Background}
Commonly used statistical models for payment card fraud detection include unsupervised and supervised methods \cite{bolton2002, ABDALLAH201690}. Whereas most systems employ supervised techniques, deciding among the two depends on business restrictions, mainly on whether historical labels exist or not. This paper assumes the availability of labels and, thus, uses supervised classification.


Without profiles, traditional tree-based models can only analyze the current transaction to inform predictions, ignoring the users' historical behavior. Profiles, in the context of this paper, are arithmetic aggregations of the data by a particular field and over a specified time window such as counts (e.g., number of transactions per card in the last 24 hours), sums (e.g., total amount spent per card in the last hour), ratios (e.g., average value of purchases for a merchant in the last week), or count distinct (e.g., number of unique cards used
per terminal in the last hour). Building and managing profiles is far from trivial:
\begin{itemize}
    \item \emph{Training with profiles}. Building good profiles is work, time, and computation intensive, as, typically, there are many possible sensible aggregations. Therefore, model development with profiles is expensive, as it requires data scientists to generate new datasets, and evaluate many resulting models. This process slows down model development time and requires specialized domain knowledge.
    \item \emph{Prediction with profiles}. The use of profiles implies their real-time availability at prediction time. As a result, they need to be computed or retrieved for incoming transactions before making predictions. Fraud detection systems, however, have to respect stringent latency SLAs in production, hence, managing profiles can become an engineering bottleneck.
\end{itemize}

Deep Learning (DL) has proved successful for unstructured data \cite{LeCun2015}, including speech recognition or object detection, and, less frequently, structured data \cite{Brbisson2015ArtificialNN}. LeCun \emph{et al.} \cite{LeCun2015} described DL as \emph{representation learning}, as it receives the raw data and learns suitable representations, providing an alternative to feature engineering and profiles. 
Feedforward Networks (FNNs) are among the fundamental types of DL in \cite{Schmidhuber2015DeepLI}. Guo and Li \cite{neuraldatamining} use separate FNNs, one per feature group, to detect credit card fraud. Similar transactions are grouped and classified by an FNN trained on other similar transactions. FNNs, however, are \emph{acyclic} \cite{Schmidhuber2015DeepLI} and, hence, unable to capture historic patterns and, ultimately,
replace profiles. 

%
%
%
Recurrent Neural Networks (RNNs) are \emph{cyclic} DL architectures based on \cite{Rumelhart1986}, addressing arbitrary sequences of input patterns \cite{Schmidhuber2015DeepLI}. RNN architectures include Long Short-Term Memory (LSTM)~\cite{doi:10.1162/neco.1997.9.8.1735} and Gated Recurrent Unit (GRU) \cite{ChoMGBSB14}. Their cyclic nature allows RNNs to encode patterns on \emph{sequences} of events, i.e., learning sequence features, and provides an intriguing alternative to profiles.

Ando \emph{et al.} \cite{Ando2016DetectingFB} use LSTMs to detect fraudulent behavior in weblog data. They fit the model on an extensive sequence of events, ordered chronologically. However, our problem is, in fact, a \emph{sequence of sub-sequences} per card. 
Li \emph{et al.} \cite{DBLP:journals/corr/abs-1711-01434} treat transaction fraud as sub-sequences of events by account. The authors separate the accounts into sets according to transaction counts and train an independent GRU for each set. A limitation of this approach is that, in a production scenario, sequences are unbounded and, therefore, we cannot statically assign accounts to sets.
Particularly, it is unclear what would happen to the hidden states when an account changes sets. 
In addition, whereas in our proposed solution, the prediction depends only on the most recent recurrent state, the authors use all the intermediate states. The proposed approach would require managing history, sets, models, and states, in latency-sensitive environments.



Jurgovsky \emph{et al.} \cite{JURGOVSKY2018234} employ LSTMs to do sequence classification for fraud detection in e-commerce and face-to-face settings, integrating also feature aggregation metrics, i.e., profiles. The authors separate the data into two exclusive sets, one for each setting. In our approach, to leverage the available information, we define scorable and non-scorable instances in training: we use the first for the forward (state update) and backward (parameter update) passes and the latter only for the forward pass.
Another crucial difference is that they consider short and long sequences of 5 and 10 \emph{time-steps}, respectively.  We train with sequences of hundreds of events following the unbounded production scenario. A curious finding, in their experiments, is that the LSTMs benefit from feature engineering.

Finally, none of the above works deal with the practical problem of implementing their solutions in a production system.

\section{Fraud detection system}

The proposed fraud detection system leverages RNNs, composing: 
\begin{itemize}
    \item \emph{offline} components to do preprocessing of the data, model training, and batch prediction, as well as;
    \item \emph{online} components to generate predictions for a stream of incoming transactions under rigorous throughput and latency constraints.
\end{itemize}

These two components support the model iteration cycle and the deployment of trained models. First, the architecture and the hyperparameters of the model are adjusted offline. Then, the model is deployed in an environment that fits the production constraints for real-time systems processing transactions. Importantly, since financial \emph{transactions} arrive in real-time systems in the form of \emph{events}, throughout this paper, we use both terms interchangeably. 


\subsection{Problem formulation}

We cast the problem in a supervised learning setting. Each instance is an event denoted by a vector $\boldsymbol{x}$ labeled as fraudulent, $y=1$, or legitimate, $y=0$. More or less information can be added depending on the use-case; however, in general, we assume $\boldsymbol{x}$ to contain:

\begin{itemize}
    \item $N_n$ numerical fields $x_{n_i}$, $i=1$ to $N_n$, containing at least the amount involved in the transaction, 
    but also possibly other fields (e.g., the geo-location coordinates of the transaction or the number of items purchased);
    \item $N_c$ categorical fields 
    $x_{c_j}$, $j=1$ to $N_c$, 
    usually strings, such as the merchant category code (MCC), the merchant's name, country code, currency code, or input mode of the card data (e.g., chip, magnetic stripe, manual input, web browser);
    \item $N_t$ timestamp fields 
    $x_{t_k}$, $k=1$ to $N_t$,
    containing at least the timestamp of the transaction  but also possibly including the expiry and issuing dates of the bank card;
    \item an entity identification field, usually a unique ID of the credit or debit card involved in the transaction, 
    $x_{{id}}$.
\end{itemize}

The system produces a prediction $\hat{y} \in \left[0,1\right]$ for a subset of instances, henceforth referred to as \emph{scorable instances}, which are determined according to some business logic. The system decides to approve or block the transaction based on whether this prediction is above a certain threshold, $y_{thr}$. 

\subsection{Model architecture}
\label{subsec:modelarchitecture}
The history of events of an entity conveys essential information for classification. Similar events can have different risk levels for different entities, depending on how they compare with the event history of that entity. As an example, a transaction of \$10,000 could be legitimate if the non-fraudulent history of the user includes frequent transactions of this amount. In contrast, it may be fraudulent if the history of the user includes only low-amount transactions. 
The architecture of the system reflects this by modeling the probability of fraud for the $i^{th}$  
event from entity $k$ as conditioned on the current and previous 
events
by the same entity. 
Additionally, we assume that a decision can depend on past events of an entity through a fixed-size state vector $\boldsymbol{s}$ for that entity that encodes information from past events as follows:
%
%
\begin{align}
    P \left( y^{(i, k)} \right) &= P \left( y^{(i, k)} \mid \boldsymbol{x}^{(i, k)}, \boldsymbol{x}^{(i-1, k)}, ..., \boldsymbol{x}^{(1, k)} \right) \\
    &= P \left( y^{(i, k)} \mid \boldsymbol{x}^{(i, k)}, \boldsymbol{s}^{(i, k)} \right).
\end{align}

Importantly, the state associated with entity $k$ after its $i^{th}$  
event, denoted above by $\boldsymbol{s}^{(i, k)}$, is assumed to depend only on the state before the event and the data contained in the event. Thus, we adopt the following recursive update of
the state $\boldsymbol{s}^{(i, k)}$ to compute the model prediction, $\hat{y}^{(i, k)}$:
\begin{align}
    \boldsymbol{x}'^{(i, k)} &= f \left( \boldsymbol{x}^{(i, k)} \right) \label{data_preprocess_eq} \\
    \boldsymbol{s}^{(i, k)} &= g \left( \boldsymbol{s}^{(i-1, k)}, \boldsymbol{x}'^{(i, k)} \right) \label{recurrent_block_eq} \\
    \hat{y}^{(i, k)} &= h \left( \boldsymbol{s}^{(i, k)}, \boldsymbol{x}'^{(i, k)} \right) \label{classifier_block},
\end{align}
where:
\begin{itemize}
    \item $f$ is a \emph{feature engineering and transformation block} which extends and converts the original input vector $\boldsymbol{x}$ into $\boldsymbol{x}'$, with additional features and representing the information in a dense space to feed typical neural networks layers (e.g., fully connected layers or recurrence cells);
    \item $g$ is a \emph{recurrent block}  which determines how to update the state after an event;
    \item $h$ is a \emph{classifier block} which generates a prediction based on the current event and state.
\end{itemize}

We assume that the three blocks contain learnable parameters, which makes RNN models especially suitable for our problem. Specifically, we use GRUs as our recurrent units, where each entity is an independent sequence with its state, but sharing the learnable parameters in $f$, $g$, and $h$ across sequences.  
With GRUs as the recurrent block, the general process of updating a new state described in equation \ref{recurrent_block_eq}, becomes: 
\begin{align}
    \boldsymbol{r}^{(i, k)} &= \sigma \left( \boldsymbol{W}^{(r)}\boldsymbol{x}'^{(i, k)} + \boldsymbol{U}^{(r)}\boldsymbol{s}^{(i-1, k)} + \boldsymbol{b}^{(r)} \right) \\
    \boldsymbol{z}^{(i, k)} &= \sigma \left( \boldsymbol{W}^{(z)}\boldsymbol{x}'^{(i, k)} + \boldsymbol{U}^{(z)}\boldsymbol{s}^{(i-1, k)} + \boldsymbol{b}^{(z)} \right) \\
    \boldsymbol{s}'^{(i, k)} &= \tanh \left( \boldsymbol{W}\boldsymbol{x}'^{(i, k)} +  \boldsymbol{r}^{(i, k)} \odot \boldsymbol{U}\boldsymbol{s}^{(i-1, k)} + \boldsymbol{b} \right) \label{apply_reset_eq} \\
    \boldsymbol{s}^{(i, k)} &= \boldsymbol{z}^{(i, k)} \odot \boldsymbol{s}^{(i-1, k)} + \left( 1-\boldsymbol{z}^{(i, k)} \right) \odot \boldsymbol{s}'^{(i, k)} \label{apply_update_eq},
\end{align}
where $\boldsymbol{r}^{(i, k)}$ denotes the reset gate, $\boldsymbol{z}^{(i, k)}$ the update gate and equations \ref{apply_reset_eq} and \ref{apply_update_eq} apply them respectively. $\boldsymbol{U}^{(r)}$, $\boldsymbol{U}^{(z)}$, $\boldsymbol{W}$, $\boldsymbol{U}$ are learnable weight matrices and $\boldsymbol{b}^{(r)}$, $\boldsymbol{b}^{(z)}$, $\boldsymbol{b}$ are learnable bias vectors.


We tested various GRU-based model architectures on several datasets. While different datasets and use-cases may require small tweaks to the architecture, our general model structure is composed of three blocks, denoted by equations \ref{data_preprocess_eq}-\ref{classifier_block}.


\subsubsection{Feature engineering and transformation block}
The data transformations, denoted by block $f$ in equation \ref{data_preprocess_eq}, are depicted in Figure \ref{fig:model_architecture}. The transformations in blue are not learnable, and as such, can be precomputed before training in the offline scenario. These are described in detail in section~\ref{subsec:preprocessing}.
By contrast, the transformation in green is learnable and cannot be precomputed before training. It consists of one embedding block per categorical feature which maps integer indices to vectors in a dense space of predefined dimension. Due to memory constraints, we map only the $k=10000$ most common values for each categorical to their embedding; we map the remaining values to the same embedding.

The transformations are concatenated. The resulting feature vector is passed through a fully connected layer to reduce its dimensionality before feeding it to the recurrent block. 

\subsubsection{Recurrent block}

\begin{figure}
  \includegraphics[width=0.5\textwidth]{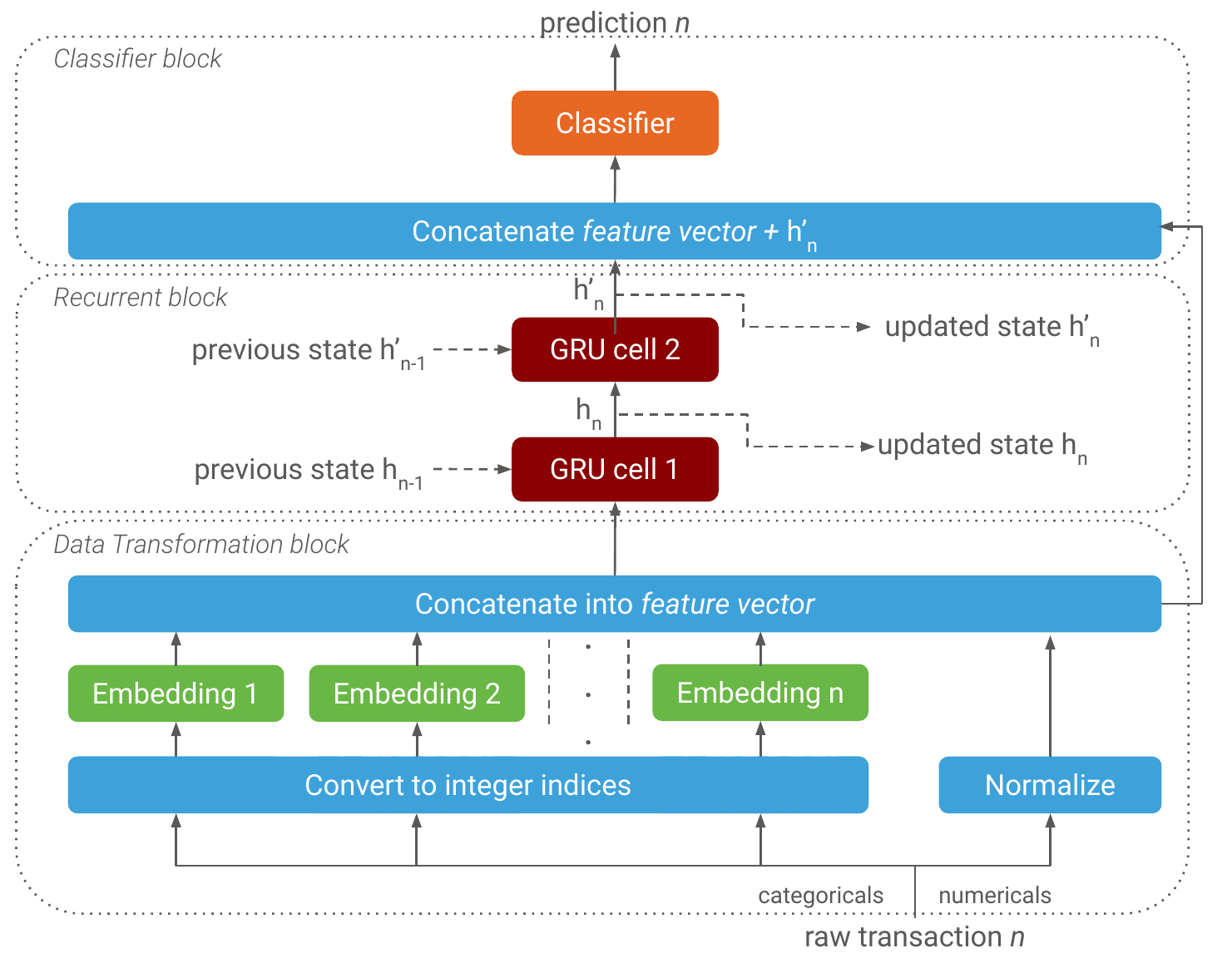}
  \caption{Illustration of the entire model architecture}
  \label{fig:model_architecture}
\end{figure}

The recurrent block is responsible for learning the consumption patterns of the cardholders and embedding this information in recurrent states. This block can be composed of a single GRU or stacked GRUs, where the output of the first becomes the input of the next. We tune the state sizes and number of GRUs with a critical constraint in mind - the storage capacity of your system. The larger the recurrent states, the larger the required database and resulting resources. As we shall see in section~\ref{subsec:streaming}, the database has as many entries per card as the number of stacked GRUs.
As a result, we carefully tune the recurrent block to maximize the fraud detection results while respecting the resource restrictions of the production system. As an example, in Figure \ref{fig:model_architecture}, the recurrent block is composed by two stacked GRUs.

\subsubsection{Classifier block}

We concatenate the output of the recurrent block with the initial feature vector so that the current prediction, $\hat{y}^{(i, k)}$, depends on $\boldsymbol{x}^{(i, k)}$ directly and through the state $\boldsymbol{s}^{(i, k)}$. This approach is not standard in RNNs, but it is comparable to a \emph{skip connection}, commonly used in Convolutional Neural Networks (CNNs) \cite{DBLP:journals/corr/LongSD14}. Hence, residual information that is useful for the current transaction, but not necessarily for future events, does not need to be stored in the hidden state. The concatenated vector passes through several fully connected layers. Eventually, it produces a final score.


%

\subsection{Offline preprocessing}
\label{subsec:preprocessing}

Training requires offline preprocessing steps that convert the data to an adequate format and implement the subset of non-learnable transformations in block $f$. To this end, we create new features and transform existing ones into appropriate numerical formats depending on their semantics, effectively transforming an original feature vector $\boldsymbol{x}$ into a processed vector $\boldsymbol{x}'$\footnote{Most of these transformations were inspired by \href{https://developers.google.com/machine-learning/data-prep/}{Google's Data Preparation and Feature Engineering in ML}}.

\subsubsection{Numerical features}
\label{subsec:numerical_preprocessing}
Transformations for numerical fields use one of two possible strategies:
\begin{enumerate}
    \item Z-scoring with outlier clipping for features with distributions that are not very skewed and with which we expect the fraud risk to vary smoothly, such as amount in US dollars:
    \begin{align}
    x_{n_i}^* &= \frac{x_i - \mu_{x_i}}{\sigma_{x_{n_i}}} \\
    x'_{n_i}  &= \max \left( \min \left( x_{n_i}^*, T_o \right), -T_o \right) 
    \end{align}
   where $\mu_{x_i}$ and $\sigma_{x_i}$ denote, respectively, the mean and standard deviation of the values of features $x_i$ in the training set, and $T_o$ is the number of standard deviations from the mean above which to consider a value to be an outlier (usually $T_o=3$);

    \item Percentile bucketing for features with multimodal distributions, or with which we do not expect the fraud risk to vary smoothly, such as latitude or longitude. Percentile bucketing amounts to creating bins between every pair of consecutive percentiles computed from the training set, and transforming feature values to the index of the bin in which they land:
    \begin{equation}
        x'_{n_i} = 
        \begin{cases}
            0, & \text{if}\ x_i < P^1_{x_i} \\
            1, & \text{if}\ P^1_{x_i} \leq x_i < P^2_{x_i} \\
            ... \\
            99, & \text{if}\ P^{99}_{x_i} \leq x_i \\
            100, & \text{if}\ x_i\ \text{has a missing or invalid value}
        \end{cases}
    \end{equation}
    where $P^k_{x_i}$ denotes the $k^\text{th}$ percentile computed over the values of feature $x_i$ in the training set. These transformed features are interpreted later as categoricals.
\end{enumerate}

\subsubsection{Categorical features}
We index each categorical feature by mapping each possible value into an integer based on the number of occurrences in the training set. For a given categorical feature, $x_{c_j}$, the $l^{th}$ most frequent value is mapped to the integer $x'_{c_j} = l-1$. All values below a certain number of occurrences map to the same integer $l_{max}$. Missing values are considered a possible value.

\subsubsection{Timestamp features}
\label{subsec:timestamp_preprocessing}
The event timestamp feature is transformed into the sine and cosine of its projection into daily, weekly, and monthly seasonality circles, i.e., a timestamp $x_{t_k}$ 
generates:
\begin{itemize}
    \item hour-of-day features $\sin (h_k)$ and $\cos(h_k)$,
    \item day-of-week features $\sin (dw_k)$ and $\cos(dw_k)$,
    \item day-of-month features $\sin (dm_k)$ and $\cos(dm_k)$,
\end{itemize}
where: 
\begin{itemize}
    \item $h_k = \text{hour\_from\_timestamp} (x_{t_k}) \cdot \frac{2 \pi}{24}$;
    \item $dw_k = \text{day\_of\_week\_from\_timestamp} (x_{t_k}) \cdot \frac{2 \pi}{7}$;
    \item $dm_k = \text{day\_of\_month\_from\_timestamp} (x_{t_k}) \cdot \frac{2 \pi}{30}$.
\end{itemize}

New features are also created by computing the difference between pairs of relevant timestamps, such as differences between:
\begin{itemize}
    \item current timestamp \emph{minus} card issue timestamp;
    \item current timestamp \emph{minus} card expiry timestamp.
\end{itemize}


We transform all of the features through the z-scoring and outlier clipping process previously described in ~\cref{subsec:numerical_preprocessing}. We do not consider yearly seasonality because the datasets used to train and evaluate models usually span a maximum of one year.

\subsubsection{Entity-based features}
\label{subsec:entitybased_preprocessing}
The entity identification field $x_{id}$  
is not mapped directly into a new feature. Instead, we group all transactions by the entity, sort each group chronologically by transaction timestamp, and then compute the difference to the immediately preceding event within the same group. The value of this feature for the $i^{th}$ 
event of entity $k$ is:

\[{x^*_{\Delta t}}^{(i, k)} = x_t^{(i, k)} - x_t^{(i-1, k)}\]
where $x_t^{(i, k)}$ denotes the timestamp of the $i^\text{th}$ event for entity $k$. This feature $x^*_{\Delta t}$ is later treated as a numerical field and transformed into its final version $x'_{\Delta t}$ through the z-scoring and outlier clipping process described in \cref{subsec:numerical_preprocessing}.
This feature is especially important because of the irregular time intervals between events. GRUs, by default, assume a constant time interval between time-steps. This irregularity conveys information that must be available to the classifier since it can significantly impact the probability of an event to be fraudulent. A representative example would be the case of ten identical transactions made over ten days or over ten seconds -- the latter pattern is more likely to represent fraudulent behavior.

A corner case is the first transaction of an entity. For practicality, the data collection process truncates all sequences at a given point in time in the past. It is impossible to tell whether a transaction was the first one for an entity or if it is the first \emph{in the dataset}. Hence, we impute the value of this feature for the first transaction in each sequence in the dataset with a manually-tuned value, usually slightly higher than the average time between consecutive transactions for the same entity (e.g., 30 days).

\subsubsection{Sequencing transactions}
Besides transforming the original feature vector $\boldsymbol{x}$ into $\boldsymbol{x}'$, we also need to transform the dataset from a format where each row is an event and the order of the rows is random, into a format where:
i) each row is a sequence of events, separated with a specific character; ii) within each sequence, the events are sorted chronologically; iii) supports quick random access of sequences to efficiently build training batches.

Additionally, as it becomes clear in \cref{subsec:training,subsec:batch_inference}, the training, and batch inference processes have different requirements in the order in which the sequences should appear, so we generate these data subsets separately in slightly different ways.

\subsubsection{Implementation}

The entire preprocessing pipeline is 
a sequence of Apache Spark jobs. These jobs take an entire dataset, usually comprising a period of up to one year, and:
\begin{enumerate}
\item create time features (section~\ref{subsec:timestamp_preprocessing}) and entity-related features (section~\ref{subsec:entitybased_preprocessing});
\item fit and store the transformer objects for normalization, bucketing and indexing (as described in sections~\ref{subsec:numerical_preprocessing} and \ref{subsec:timestamp_preprocessing}) based on the training period;
\item load and apply all transformers to the training, validation and test datasets;
\item turn a list of events into a list of chronologically-sorted sequences of transactions using \texttt{groupBy} to group events by the same entity and \texttt{map(sort\_by\_event\_timestamp)} to sort the events within each group chronologically;
\item for each of the intended output subsets (training, validation, evaluation), filter out the sequences that do not contain any event in the period of interest;
\item store subsets in databases for quick CPU access.
\end{enumerate}

\subsection{Training}
\label{subsec:training}

\begin{figure*}[ht]
  \includegraphics[width=0.7\textwidth]{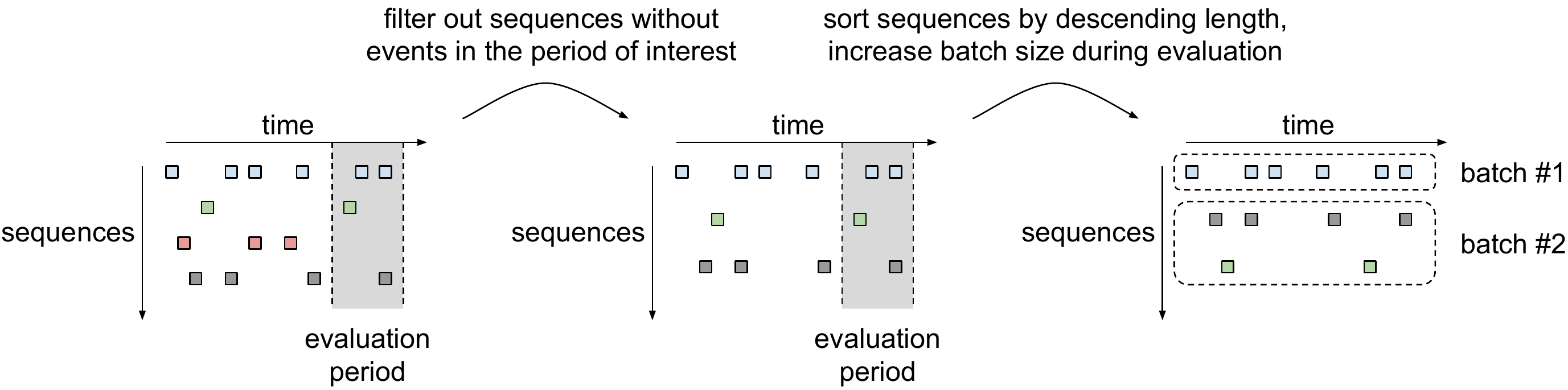}
  \caption{Steps involved in running batch inference.}
  \label{fig:batch_inference}
\end{figure*}

In fraud detection, we can train and evaluate our models through offline batch processing. However, strict business SLAs require that, in production, transactions must be scored in real-time, changing the paradigm into stream processing.  There are implications when using batch inference to obtain a realistic, unbiased estimate of the model's performance, as discussed in section \ref{subsec:batch_inference}. Furthermore, given that we are working with time-series data, we must ensure that the training procedure does not give an unrealistic advantage to the model trained with historical data. More concretely, the model must not be able to extract information in any way from the future. Hence, our preprocessing stage ensures that each card's transactions are saved in the database by chronological order (and fed to the model in that same order), and standard hyper-parameter search techniques such as cross-validation are not employed. Below, we shall discuss the details of our approach from the engineering and machine learning perspectives.

\subsubsection{Data engineering details}

Given the size of the data we are processing (each dataset occupies several TBs of disk space), I/O is a bottleneck of our training.
%
Hence, we made several optimizations in order to: i) speed-up the access and storage of data; ii) efficiently build training batches for the classification problem we are solving.

Firstly, we use the embedded key-value store LMDB \cite{lmdb, chu2011mdb} for its read performance to store the training, validation, and test datasets (each subset as a separate database). Instead of building random training batches of cards through several separate random reads, for every batch, we access an index of the database at random and read sequentially from there on batch size number of entries. This way, we ensure that the model consistently sees different batches during training in an efficient manner. We also employ a multi-processing strategy where various CPU workers are continuously and concurrently reading batches from the database and inserting them in a queue to be later consumed by the GPU.

The last optimization is due to the extremely imbalanced problem we are working with (cf. Table~\ref{tab:datasets} for dataset characterization). We want to enforce that, at every batch, the model sees some fraudulent transactions. Otherwise, back-propagating the gradients derived from a batch with no fraud would result in a model that simply predicts all transactions to be non-fraudulent. As a result, we must do aggressive fraud sampling, which, if implemented naively, is an exceptionally inefficient process.
Consequently, our tool builds two distinct databases, one of the cards with no fraudulent transactions, and another with cards containing at least one fraudulent transaction. Since the fraudulent database is much smaller than the non-fraudulent one, we can keep it in memory (RAM) hence making the fraudulent reads, which occur for every batch, even faster. For the validation dataset, we build a new database of a random and representative sample of cards from the validation period (typically a month) arranged in descending order in the number of transactions. This allows us to quickly find the batch size that maximizes the GPU's usage, and there is no problem with having the model evaluate the transactions always in the same order.


\subsubsection{Machine learning details}

In a production scenario, an RNN continuously updates its recurrent state from the very first transaction of a card until its most recent one. In the offline batch scenario, forcing a model to classify the entire transactional history of a card is undesirable for two reasons. 
Firstly, the transactional history of some cards may be so long that it surpasses the GPU memory, whereas others may have only a few associated transactions.
This disparity 
causes very irregular batches
, and an inefficient usage of the GPU memory.
Secondly, domain knowledge dictates that one does not need the entire history of a card to decide 
if the current transaction is fraudulent. Because of this, we implement a cutoff on the number of previous transactions of a card. This cutoff must be empirically set, depending on the dataset and use-case, to establish a good compromise between GPU utilization and a sufficient history for the model to distinguish between the user's normal behavior and a fraudulent one. Empirically, we found that in the payment processor domain, a range from 100 to 400 previous transactions seems to establish this compromise. Consequently, our batches have a fixed size in the number of transactions resulting from the number of cards (batch size) multiplied by the cutoff length. The motivation behind using long sequences is to mimic, as much as possible, the unbounded number of transactions, per card, in production.  

Even though clients typically provide, in their historical data, transactions from all of their channels and use-cases, in many cases, they only want us to score transactions from a specific subgroup. As an example, transactions can be either card-not-present (CNP) for online transactions or card-present (CP) for transactions in physical locations,
and a client may ask us to build a model that has the sole purpose of blocking CNP transactions. Although the model is only scoring CNP transactions, it can still extract valuable information from CP transactions. More concretely, imagine a sequence of successive cash withdrawals followed by online purchases done in a short period. The model would be further inclined to detect this suspicious behavior if it had seen the CP transactions before the CNP ones. Hence, we embed this information directly into the training procedure through the use of scorable and non-scorable instances. A transaction that the model needs to learn how to classify goes through the typical forward pass, followed by the backpropagation of the respective gradients. For a non-scorable instance, however, a forward pass is done, but the backward pass is not. As a result, with the forward pass, the recurrent state of the card is updated with new information. Still, the model does not learn how to classify the non-scorable instances, focusing solely on the target use-cases.

We split the dataset into train, validation, and test subsets. Since this is a binary classification problem, we use the binary cross-entropy loss to train our models. 
Given the data's size, an epoch is not an entire pass through the data.
Instead, an epoch is a random sample, with replacement, of, approximately, 10\% non-fraudulent cards and all fraudulent cards in the training data. 

\subsection{Batch inference}
\label{subsec:batch_inference}

Obtaining accurate, unbiased estimates of the performance of these models requires us to compute predictions for all events in a full period of data in the offline setting. It is similar to the validation process executed periodically during training as described above, with some crucial differences:
\begin{enumerate}
    \item truncating sequences to the last $m$ 
    events is no longer feasible as we want to generate predictions for all transactions in a time window;
    \item ensuring sequence randomness is no longer needed, since we evaluate all sequences. 
\end{enumerate}

Taking these two differences into account, we first filter out sequences that do not contain any event in the time period. Then, we sort the list of sequences by descending sequence length. This allows each batch to be composed of similarly-sized sequences, reducing the number of padded events. Finally, we adapt the batch size as we go over the data; we start with a small batch of lengthy sequences and progressively increase the batch size as the length of the sequences decreases. Thus, the number of events in each batch remains similar, keeping the GPU memory usage stable (and high) over time. \Cref{fig:batch_inference} illustrates this process; in this example, the batch size increases from one sequence to two sequences, but the total number of events in each batch is the same, as desired.

\subsection{Streaming inference}
\label{subsec:streaming}
Feedzai's main product, \emph{Pulse}, is a streaming engine used to detect fraud in real-time. Our product works directly in the hot path of approving a financial transaction. As a consequence, we have very tight latency SLAs (i.e., time to process one event). Depending on the use-case and the client, our system may take at most 200ms on the 99.999th percentile of the latencies distribution to process a transaction. This time includes several components, not all directly related with scoring a transaction, e.g., preprocessing and enriching data, or applying actions according to predefined business rules.
For this reason, we need an extremely efficient streaming inference process, where we adapt the way we provide the history of each card to the model, described next. 

\subsubsection{Implementation Overview}
\begin{figure}
  \includegraphics[width=0.5\textwidth]{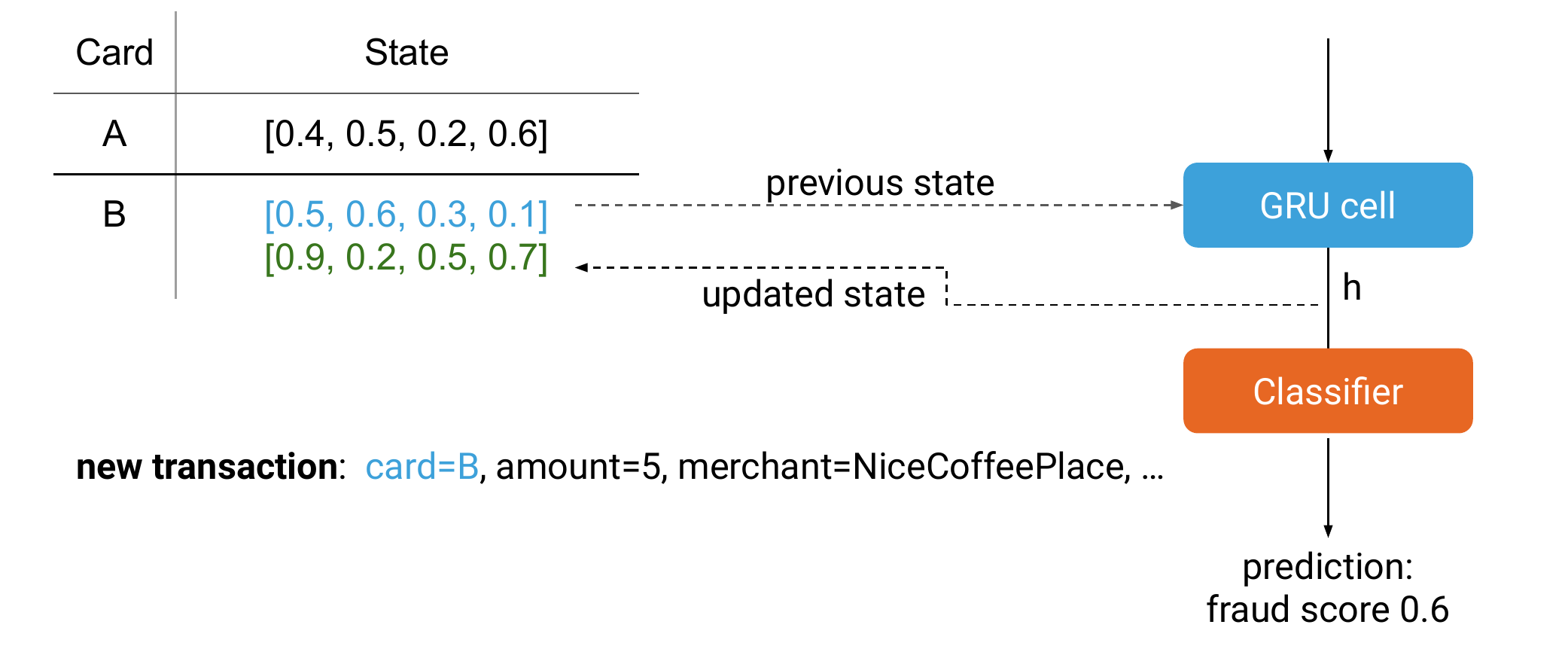}
  \caption{RNN State Update - Streaming}
  \label{fig:rnn}
\end{figure}

Previously, in a batch inference setting, sequences were explicitly ordered to maximize GPU usage, and minimize inference time. In streaming, this is no longer possible, since transactions have to be scored by the order in which they arrive to the system.
In this setting, the sequence information is encoded in the recurrent states of the model and saved externally into a key-value store. As illustrated in Figure~\ref{fig:rnn}, we store a \emph{state} for each card, and whenever a new transaction arrives we:
\begin{enumerate}
    \item Fetch the current state for the given card identifier;
    \item Feed the current transaction and the state we just fetched to a GRU cell, yielding the new state for this card number and the score for the respective transaction;
    \item Update the state in the key-value store for the given card.
\end{enumerate}

The state captures the history of transactions of a given card. 
With this, we ensure that the time and space complexity of scoring each transaction is constant, regardless of it being the first appearance or the millionth one for that card. A new card without any history has its state values all initialized to 0.0. 

\subsubsection{Storing and Accessing State}
A key-value store stores the state of each card seen by the system.  Intuitively, the key-value store grows linearly on the number of cards, as we have one entry per card. Each entry size depends on the number of GRU layers and the size of the GRUs' states. Typically, we use relatively small GRUs to minimize state size pressure. Besides, empirical experiments have shown that recurrent states with more than 128 units do not improve results and worsen training times. 

The sheer number of distinct cards for most of our clients makes it unfeasible to have all these states in memory.
%
To circumvent this, we use HaloDB \cite{halodb}, a high-performance, JVM-based, embedded key-value store. HaloDB is empowered by LSM-trees~\cite{ONeilCGO96} and an efficient garbage collection process to keep write amplification low. We use this embedded key-value store 
for various reasons:
\begin{itemize}
    \item We avoid the network overhead of fetching this state from an external machine.
    \item Reading a key is close-enough to RAM performance since HaloDB keeps an index in memory for the keys stored.
    \item By design, LSM-trees frequently checkpoint their content to disk (when flushing the \emph{memtable} contents to immutable \emph{sstable} files~\cite{ONeilCGO96}), making it easier to recover from failures.
\end{itemize}

Instead of an embedded key-value store, we could have chosen 
an external database like Redis or Cassandra. Yet, this would make our product more complex (with one extra component to operate), and slower due to the cost of network I/O. One could argue that keeping the state embedded  can be a bottleneck to achieve the system's scalability and availability. However, Feedzai designed its product to achieve this by partitioning the load into several processing instances; and by having hot-standby replicas, respectively.

\subsubsection{Additional Optimizations}
To minimize latency further, especially on the high-percentiles, we avoid writing to HaloDB synchronously (i.e., during the critical path of scoring a transaction). To this end, we also use a small Least Recently Used (LRU) cache to keep a copy of the most recent updated states in memory (RAM). Hence, whenever we need to update the state of a given card, we first write this state to the cache, and then asynchronously write the same state to our key-value store persisted on disk. 
Whenever a new transaction arrives, we first attempt to read it from the cache and, if not present, read it from disk. Please note that this LRU cache can be relatively small, as we only need to ensure that our asynchronous write to the key-value store finishes before evicting the item from the cache. From our benchmarks, a write on HaloDB takes 0.1 ms on average (even on HDD), and around 400 ms on the 99.999\%. Thus, we typically size the cache to hold some dozens of seconds of data to cope with hiccups on high throughput scenarios. 
The overall architecture can be visualized in Figure~\ref{fig:streaming-interleaved}.

\subsubsection{Expiring States}
Clearly, for some of our clients, we cannot keep states for all cards seen, forever, even when only writing this state to disk. 
To cope with this, we asynchronously expire old entities of our key-value store, periodically. This expiration happens whenever we reach some threshold in time or usage space. 



\section{Experiments}
\label{sec:validation}
To validate our approach, we ran experiments with proprietary large-scale datasets. We focus on two main goals: i) beating a strong baseline in terms of model performance and; ii) ensuring that the solution is usable in a streaming production scenario, at significant throughputs and under tight latency constraints.

\subsection{Datasets}

We use two datasets to run experiments. Both are real datasets from two major European financial institutions, processing card-present (CP) and card-not-present (CNP) transactions. Although we cannot disclose these datasets due to privacy and contractual reasons, Table~\ref{tab:datasets} shows some relevant numbers to characterize them.

\begin{table}[h]
\begin{tabular}{lll}
Dataset                                        & A    & B    \\\hline
Total number of transactions                   & 1B & 4B \\
Total number of cards (entities) involved      & 76M & 65M \\
Average number of transactions per card        & 7 & 61 \\
Ratio of fraudulent to legitimate transactions & 1:200 & 1:7000 \\
Number of raw categorical features             & 15 & 53 \\
Number of raw numerical features               & 2 & 4 \\
Number of raw time-related features            & 2 & 2 \\
Time period (months)                              & 7  & 10 
\end{tabular}
\caption{Characterization of datasets used.}
\label{tab:datasets}
\end{table}

\subsection{Fraud detection results}

\subsubsection{Setup}

We split both datasets into test (1 month), validation (1 month), and training (remaining months) sets.
The splits respect the chronological order of our data. Hence, the training sets precede the validation sets, and the validation precedes the test sets. 

We evaluate our models on the  business metrics decided by both the individual clients and Feedzai. We trained 10 models for each dataset.  
For dataset A, we found that two stacked GRUs, the first with recurrent states of 128 units and the second with 64, obtained the best results. In dataset B, the number of GRUs and state size was almost irrelevant, 
and so we use just one GRU with a recurrent state size of 64 units for engineering reasons. 

We found the sampling factor and history length parameters to influence model's performance the most. For both datasets, generating training batches with 95\% of non-fraudulent cards and the remaining 5\% with cards that have at least one fraudulent transaction seemed to work the best. Model weights were updated using Adam optimizer with the default 0.001 learning rate. We used patience of 20 epochs for early stopping, and after 10 epochs without improvement, we reduce the learning rate by a factor of 10.

For our baselines, we considered production-ready LightGBM models, one per dataset, each using around 100 profiles on top of the raw features. These models are the result of considerable domain knowledge and expertise, making them solid baselines to compare with our novel GRU-based approach, which uses only raw features.


\subsubsection{Discussion}

The results from Tables~\ref{tab:fraud-performance-A} and \ref{tab:fraud-performance-B} show that our GRU-based model without profiles outperformed the LightGBM model with profiles in the majority of the metrics\footnote{\label{approx_footnote}Due to privacy reasons only relative values can be disclosed.}. 
The only exception was in the false positive rate (FP-Rate) for dataset A (Table~\ref{tab:fraud-performance-A}), where it obtained around 0.25\% points less. The RNN model, however, convincingly outperformed the baseline in terms of recall, with 10\% points more, and 1.6\% more money recall (\$Recall) for the same dataset. We highlight that the small improvement in money recall equates to almost 1 million more euros saved\footnoteref{approx_footnote}, the value obtained by subtracting the money true positive values (\%TP). 
For dataset B, the RNN beat the LightGBM baseline in every metric, with almost 3\% points more recall and 6\% points more card recall. The 2.7\% points improvement on money recall equates to more than 8 million euros in fraudulent money blocked\footnoteref{approx_footnote}.

The most sensitive hyper-parameter was the historical sequence length (which decides how many past transactions the model should consider). For dataset A, a cutoff length greater than 150 previous transactions did not yield improvements, whereas, for dataset B,
the best cutoff length was 400 transactions.
The reason for this discrepancy resides in each dataset use-case.
Dataset A comprises only CNP transactions, while dataset B has both CNP and CP transactions. CNP transactions are riskier, as they do not require the cardholders' presence as in in CP transactions. CP transactions are also more frequent, and thus, the model trained on dataset B needed a higher number of past transactions to obtain the best results.

\subsubsection{GRUs vs LSTMs}
As an additional experiment, we measure the training stability of our GRU-based approach and compare its performance to LSTMs. We fixed a model architecture for both datasets, and only vary the recurrent model, GRU or LSTM, within our architecture's recurrent block. We trained three models of each architecture per dataset (A and B). For dataset A, GRUs obtained 0.23\% points less recall, on average, than LSTMs, with a smaller standard deviation of 0.32\% points when compared to LSTMs with 0.71\% points. Interestingly, on dataset B, GRUs got 1\% point more recall, on average, when compared to LSTMs with a similar standard deviation (0.44\% points GRU vs. 0.46\% points LSTM).

In our experiments, and for the respective target metrics of each dataset, LSTMs did not convincingly beat GRUs. Since LSTMs are more complex and have additional learnable parameters, GRUs are better to operate in production. Namely, as we manage only one recurrent state instead of two, our deployment is simplified without compromising model performance.

\begin{table}
\begin{tabular}{lllll}
Model & Recall & FP-Rate & \$Recall & \$TP \\
\hline
RNN w/o profiles & \textbf{+9.7\%} & +0.34\% & \textbf{+1.54\%} & $\approx$\textbf{+1M\euro{}} \\
\end{tabular}
\caption{Dataset A: Delta between RNN and LightGBM results, for 15\% precision, as requested by the client.}


\label{tab:fraud-performance-A}
\end{table}

\begin{table}
\begin{tabular}{llllll}
Model & Recall & \$Recall & \$TP & Card recall  \\
\hline
RNN w/o profiles & \textbf{+3.2\%} & \textbf{+2.4\%} & $\approx$\textbf{+8M\euro{}} & \textbf{+5.9\%}  \\
\end{tabular}
\caption{Dataset B: Delta between RNN and LightGBM results, for 5000 card alerts/day, as requested by the client.}
\label{tab:fraud-performance-B}
\end{table}

\subsection{Engineering performance results}
\label{subsec:engineering_experiments}

\subsubsection{Setup}

Our experiments assess the efficiency of our system concerning its latency.
To score a transaction, we used TensorFlow Serving (TFX) v.1.13.0 to host our model.
TFX is deployed as a Docker image, on the same machine as \emph{Pulse} (Feedzai's streaming engine), to avoid any network communication overhead.  
We used a machine with 32 Intel Xeon CPU E5-2680 v3 @ 2.50GHz and 50GB of RAM. To optimize performance, we compiled the TFX Docker Image to our machine architecture.
We use standard CPUs since GPUs are not particularly worthy in a streaming setting where we need to score each instance individually. We used dataset A and a model with two stacked GRUs with unit sizes of 128 and 64.

To measure the latency for each event, we used an external injector, pushing events to \emph{Pulse} at a constant rate of 500 events per second. Note that, while 500 events might appear a low value for some financial settings, it is also our usual threshold for partitioning the load into several machines
, which means that for scenarios demanding, e.g., up to 4000 events per second, we use 8 servers in parallel.
To prove the system's stability, we ran the experiment for one whole day, measuring the time spent by the system from different probes: 
i) the writing time to disk, done asynchronously; ii) the reading time from the cache, or disk (if the element is not present in the cache); and iii) the total time to make a prediction. The latter measures the pipeline in its entirety, from the moment a transaction arrives to the moment a score is output.
The results are shown in Table~\ref{tab:eng-performance}. 
\begin{table}[h]
\begin{tabular}{lrrrrr}
Probe (ms) & mean &99\% &99.9\% &99.99\% &99.999\% \\
\hline
Write disk (async)	&0.05	&0.06	&0.37	&62.37	&398.70 \\
Read (cache or disk) 	&0.01	&0.01	&0.10	&0.50	&3.13 \\
Total prediction time 	&4.06	&10.47	&42.82	&75.90	&126.66 \\
\end{tabular}
\caption{Engineering performance - 500 events/s for 1 day.}
\label{tab:eng-performance}
\end{table}
\subsubsection{Discussion}

As described in section~\ref{subsec:streaming}, our models score transactions in the hot path of mission-critical financial systems, which are incredibly latency-sensitive. Depending on the use-case and  client, our system may take at most 200 ms on the 99.999th percentile to fully process a transaction. Hence, we often engineer our systems for high percentiles, at the expense of the average case.

Table~\ref{tab:eng-performance} shows that we can couple RNNs to Feedzai's Streaming Engine to make predictions in production while adhering to our most common SLAs.
Still, the standard Feedzai setup is faster, on average, and in most percentiles, than our RNN setting. On average, a typical Feedzai setup takes around 2 to 3 ms to process a transaction. Most of Feedzai's production settings use tree-based models such as Random Forests or LightGBM, which are more straightforward and generally faster. However, on these setups, we also need to compute profiles, and in some cases, fetch them from external systems. The latter is, in fact, the primary source of our product's latency in high percentiles. As a result, we observed that RNNs without profiles are faster than some of our usual setups for high percentiles (>$99.999\%$).


Note that using a cache to avoid synchronous writes to our embedded key-value storage is paramount to our success. Without it, we would worsen the results considerably in high percentiles, and breach our target SLAs. We used Yahoo's HaloDB key-value store for the simplicity of its API, easy integration with the JVM ecosystem, and its performance. Further internal tests have shown that Facebook's RocksDB~\cite{rocksdb} could be an even better choice, given its proven stability and high adoption in other production systems. 

We tried to optimize our model using Tensorflow's performance best practices, 
including
\emph{Pruning} and \emph{Quantizating} \cite{tf-optimization}. However, these did not provide any relevant latency improvements for our models and setup. Surprisingly, compiling Tensorflow to the right CPU architecture gave us a 5x reduction in inference latency. 

\section{Conclusions}
We present an RNN framework for detecting financial fraud in real-time, describing all of its stages from preprocessing, model architecture, offline training, to production. 
The cyclic nature of RNNs allows us to efficiently capture patterns on sequences of events, without the need to use intermediate feature engineering (profiles). Unlike most DL setups in structured data, our RNN model outperforms our best tree-based learners in the various metrics evaluated. In addition, it does so while complying to millisecond-level latencies. Key to our success, is the way we treat sequences and the whole architecture of our system.
To support predictions in streaming, where transactions arrive with an unexpected order, we treat sequences as interleaved sub-sequences with irregular time between transactions. 
We also assume the history of transactions to be unbounded, but compressed as a GRU state of fixed size.

Since we do not need profiles, this setup has several advantages. From an ML perspective, we avoid the expensive task of building and iterating profiles. From an engineering perspective, we avoid the costs and complex logic to compute, store, update and access profiles, simplifying our deployments.
 Lastly, besides showing how our RNN model can be used in mission-critical low-latency production systems, we also provide valuable lessons on how to process the data, implement an  efficient training scheme, and design the model's architecture to support predictions in streaming.

As next steps, we still want to know how the inclusion of profiles in our model can provide overall gains in performance, and what type of profiles yield the most improvements. The latter is essential to conclude what type of information RNNs can intrinsically encapsulate, and build impactful profile features. 
%
%
Finally, we also want to change the GRU architecture to take the irregular spaced time intervals between transactions into consideration. 
Hence, instead of feeding it as a feature, hoping that the model can learn to interpret it correctly, we would leverage this information and directly embed it in the mechanism that updates the recurrent states. 

\bibliographystyle{ACM-Reference-Format}
\bibliography{rnn-paper}

\end{document}